\documentclass[sigconf, nonacm]{acmart}

 \usepackage{graphicx} 
 \usepackage{comment}

\usepackage{xspace}
\usepackage{enumerate}
\makeatletter
\newcommand{\myitem}[1]{%
    \item[#1]\protected@edef\@currentlabel{#1}%
}
\makeatother

\usepackage{enumitem}
\setlist[itemize]{leftmargin=*}

\usepackage{fancybox}


\usepackage{multirow}
\usepackage{geometry}
\usepackage{makecell}
\usepackage{listings}

\usepackage{pgf-pie}
\usepackage{pdflscape}
\usepackage{pgfplots}
\pgfplotsset{compat=1.18}
\usepgfplotslibrary{statistics}
\usepackage{tikz}
\usepackage{sfmath}
\tikzset{
  barlabels/.style={font=\footnotesize\sffamily},
  declare function={
    barheight=5pt;
  }
}

\usepackage{colortbl}
\usepackage{subcaption}
\usepackage{units}

\usepackage{color,soul}
\usepackage{geometry}

\definecolor{javared}{rgb}{0.6,0,0} 
\definecolor{javagreen}{rgb}{0.25,0.5,0.35} 
\definecolor{javapurple}{rgb}{0.5,0,0.35} 
\definecolor{javadocblue}{rgb}{0.25,0.35,0.75} 

\newcommand{\policy}{U-MAP\xspace}
\newcommand{\policies}{U-MAPs\xspace}
\newcommand{\policiesfull}{User-Managed Access Control Policies\xspace}
\newcommand{\va}{VA\xspace}
\newcommand{\vas}{VAs\xspace}
\newcommand{\vasfull}{Virtual Assistants\xspace}

\newcommand{\questionone}{\textbf{RQ1}: \emph{Can \vas handle \policies effectively from a security perspective?}\xspace} 
   
\newcommand{\questiontwo}{\textbf{RQ2}: \emph{What recommendations can be given to future \vas to effectively and efficiently handle \policies?}\xspace}

\definecolor{gray}{rgb}{0.4,0.4,0.4}
\definecolor{darkblue}{rgb}{0.0,0.0,0.6}
\definecolor{cyan}{rgb}{0.0,0.6,0.6}
\definecolor{yellow-pastel}{rgb}{255,242,204}
\definecolor{green-pastel}{rgb}{213,232,212}
\definecolor{blue-pastel}{rgb}{218,232,252}
\definecolor{red-pastel}{rgb}{248,206,204}
\definecolor{green-dark}{RGB}{0,100,0}

\newcommand\checkmarknew[1][]{%
  \tikz[scale=0.4,#1]{\fill(0,.35) -- (.25,0) -- (1,.7) -- (.25,.15) -- cycle;}%
}

\newcommand{\FALSE}{
\textbf{\textcolor{red}{$\times$}}
}

\newcommand{\TRUE}{
\checkmarknew[green-dark]
}

\title[Exploring the Evaluation of User-Managed Access Control Policies by AI Virtual Assistants]
{"\emph{I Apologize For Not Understanding Your Policy}": \\
Exploring the Specification and Evaluation of  
\policiesfull 
by
AI
Virtual Assistants}

\author{Jennifer Mondragon}
\email{jmondragon6@islander.tamucc.edu}
\author{Carlos Rubio-Medrano}
\email{carlos.rubiomedrano@tamucc.edu}
\orcid{0000-0001-8931-6412}
\affiliation{%
  \institution{Texas A\&M University- Corpus Christi}
  \city{Corpus Christi}
  \state{Texas}
  \country{USA}
}

\author{Gael Cruz}
\email{cruzg29@gator.uhd.edu}
\author{Dvijesh Shastri}
\email{shastrid@uhd.edu}
\affiliation{%
  \institution{University of Houston - Downtown}
  \city{Houston}
  \state{Texas}
  \country{USA}
}

\begin{abstract}

The rapid evolution of Artificial Intelligence (AI)-based \vasfull (\vas) e.g., Google Gemini, ChatGPT, Microsoft Copilot, and High-Flyer Deepseek has 
turned them into 
convenient interfaces for managing emerging technologies such as Smart Homes, Smart Cars, Electronic Health Records, 
by means of explicit commands, e.g., prompts, which can be even launched via voice, thus providing a very convenient interface for end-users.  
However, the proper specification and evaluation of 
\policiesfull (\policies), 
the rules issued and managed by end-users to govern access to sensitive data and device functionality - within these \vas presents significant challenges, since such a process is crucial for preventing security vulnerabilities and privacy leaks without impacting user experience. 
This study provides an initial exploratory investigation on whether current publicly-available \vas can manage 
\policies
effectively across differing scenarios. By conducting unstructured to structured tests, we evaluated the comprehension of such \vas, revealing a lack of understanding in varying 
\policy
approaches. Our research not only identifies key limitations, but offers valuable insights into how \vas can be further improved to manage complex authorization rules and adapt to dynamic changes.

\end{abstract}

\begin{document}

\maketitle

\section{Introduction}\label{sect-intro}

\vasfull (\vas) 
powered by Large Language Models (LLMs) \cite{chang2024survey, kasneci2023chatgpt, mbakwe2023chatgpt} are becoming trendy, as they provide a highly convenient human interface, suitable for many different scenarios and application domains such as 
Smart Homes, in which they aid in controlling smart devices (TVs, Lights, Locks)~\cite{edu2020smart}, Smart Cars, in which they can be useful for giving driving directions and controlling functionality via voice commands~\cite{guanetti2018control}, as well as in Electronic Health Records (EHR), as they can provide patient information in an expedite and efficient manner during mission-critical operations such as surgery or first-response emergencies~\cite{kumah2018electronic}.
Not surprisingly, several different commercial products are already in the market, either in a \emph{hardware-based} mode, i.e., Amazon Alexa and Google Nest, as well as in an \emph{online-based} approach, i.e., Apple Siri, ChatGPT, Google Gemini, Microsoft Copilot, etc. As of today, several other companies are actively working towards providing dedicated, efficient \vas for a variety of application domains~\cite{VASupreme2024}. 

In such a context, the management of 
\policiesfull
(\policies)~\cite{IEEEexample:NIST_AC}, i.e., the rules governing access to \emph{sensitive} data and functionality within computer systems, may be required in \va scenarios~\cite{Fortune2024}. As an example, the correct specification, evaluation, and enforcement of \policies may be crucial to mediate \emph{who} is allowed to control devices (Smart Homes), 
\emph{who} can give directions and change car settings (Smart Cars), and to mediate \emph{who} can access private patient information in (EHRs). 
In these scenarios, not handling \policies correctly can have serious consequences, e.g., thieves controlling a Smart Lock guarding the main door (Smart Homes), kids altering the course of action of a car and causing an accident (Smart Cars), and a surgeon missing important allergy information on a patient, causing 
unnecessary complications (EHRs).   

However, 
despite the excitement of the possibilities of \vas for improving human-computer interactions, and the many different solutions that are becoming commercially available, 
it is not clear if publicly-available, general-purpose \vas can effectively and efficiently handle \policies. More specifically, it is still unclear if 
the management of \policies via \vas correctly assigns authorization privileges/rights, a.k.a., \emph{permissions}, to protected resources, e.g., Smart Home devices and Smart Car functionality, thus potentially avoiding the introduction of security vulnerabilities otherwise, which could be exploited by malicious third parties to compromise the security of such systems. 
In addition, the current landscape of \vas provides no insights on whether 
the management of \policies is convenient to humans, e.g., it avoids difficult interactions, delays in response, 
etc., while still preserving the security properties just discussed.  

In order to address these concerns, this paper presents the very first 
exploratory investigation on whether current publicly-available and highly-popular \vas, namely, 
ChatGPT 
(Version GPT-4o),
Google Gemini
(Version 2024.09.04), 
Microsoft Copilot
(Version 10.28), and
High-Flyer Deepseek
(Version 2025.01.20)
can manage \policies effectively across differing scenarios. Specifically, our study seeks to answer the following research questions:

\begin{itemize}
    \item \questionone
    \item \questiontwo
\end{itemize}

%
Overall, 
this paper provides the following contributions:

\begin{itemize}
    \item As a part of Sec.~\ref{sect-methodology}, we present a series of specification formats developed to better communicate \policies to \vas, e.g., using plain natural language versus more \emph{structured} formats, detailing their syntactic and semantic contents, as well as their theoretical background.  
    \item Also, in Sec.~\ref{sect-results}, we present the results of an exploratory study assessing the performance of four publicly-available \vas when handling \policies. Our results indicate that the \vas under study exhibit varying degrees of proficiency depending on the format used to communicate each \policy, e.g., certain \vas demonstrate strong performance with structured \policies, but show limitations with natural language inputs. Conversely, other \vas perform well across a range of \policy formats, though specific areas require refinement.
    \item Finally, in Sec.~\ref{sect-results}, we provide a series of recommendations for enhancing the performance of \vas when managing \policies. As an example, due to struggles with unstructured \policies, integrating advanced natural language processing techniques could refine and assist in contextual understanding. Enhanced contextual understanding could empower \vas to better navigate inputs, thus improving their versatility and reliability.
\end{itemize}

This paper is organized as follows: we start in Sec.~\ref{sect-back-related} by reviewing some relevant background and related work. Then, we dive into the problem statement we address in this paper in Sec.~\ref{sect-problem}. We then move on to present the methodology we follow for our study in Sec.~\ref{sect-methodology}, and presents its results and subsequent recommendations in Sec.~\ref{sect-results}. 
Sec.~\ref{sect-limitations} discusses the limitations of our work. 
Finally, Sec.~\ref{sect-conclusions} concludes this paper with some interesting topics for future work.

\section{Background and Related Work}\label{sect-back-related}


We start by presenting some basic background on topics that we will further explore later. In Sec.~\ref{subsect-back-ai-vas} we explore the uprising of \vas as well as latest developments relating LLMs and security. Also, in Sec.~\ref{subsect-back-umaps}, we explore a definition of \policies and their relevance for handling emerging technologies. 

\subsection{LLM-based Virtual Assistants}\label{subsect-back-ai-vas}



Due to their remarkable performance, Large Language Models (LLMs) are rapidly gaining traction across a multitude of diverse domains, from finance and health care to education and software development  ~\cite{chang2024survey,chen2021evaluating, kasneci2023chatgpt, mbakwe2023chatgpt, sun2023short, yao2023benchmarking}. 
In particular, 
\vasfull (\vas) leverage LLMs as their foundational model for the processing and retrieval of general and domain-specific knowledge, and can be further augmented with additional data processing and storage capabilities by pairing their LLM-based backend with traditional databases, ontologies, knowledge graphs, etc. \cite{dong2023towards}, thus making them ideal for convenient user interfaces, through which emerging smart technologies such as Smart Homes, Smart Cars, and EHRs, can be operated naturally and intuitively. 

However, despite the growing interest and emerging applications, the use of
LLMs 
as a back-end module for \vas 
can pose security vulnerabilities, leading to undesirable outcomes for the users of these technologies. As an example, 
Yao et al. categorized LLMs's security and privacy vulnerabilities into five categories: hardware-level attacks, OS-level attacks, software-level attacks, network-level attacks, and user-level attacks ~\cite{ yao2024survey}. Pearce et al. investigated GitHub Copilot's security vulnerability in computer code generation and reported that 40\% of the 1,689 programs generated by Copilot in their experiment introduced security vulnerability  ~\cite{pearce2022asleep}. Iqbal et al. discussed security, privacy, and safety challenges related to LLMs integration with third-party plugins. The challenges include injecting malicious descriptions, diverting prompts to another plugin, and stilling plugin data   ~\cite{iqbal2023llm}. Although prior research has evaluated LLMs for security vulnerabilities and data privacy, there is no systematic examination of LLMs' ability to interpret \policies necessary to operate smart technologies effectively. 

\subsection{\policiesfull}\label{subsect-back-umaps}

For the purposes of this paper, we define \policiesfull (\policies) to be the set of access control/authorization policies whose \emph{lifecycle}, i.e., their specification, update, and removal over time, is handled mostly by \emph{end-users}, who may have not received formal / informal training in cybersecurity/ access control topics \cite{Smetters2009}.
End-users may also not receive dedicated advised from cybersecurity/access control experts when it comes to handling the lifecycle of \policies, and may instead resort to a combination of domain-specific knowledge 
and 
common sense. 

The correct management of \policies may become central to the development and adoption of modern emerging technologies, which are fading away from expert-managed, \emph{centralized} approaches, e.g., a single security officer handling all access control policies within an organization, to more \emph{distributed} approaches in which end-users are given full control of the way they interact with a given technology, restricting access to both functionality and data at will.
In such a scenario, \policies are expected to provide a balance between \emph{domain-specific} settings, e.g., conveniently restricting access to functionality, and \emph{security-specifi}c settings, e.g., the collection and retrieval of data.
Table~\ref{tab:sap:sample} provides an example of a \policy handling a Smart Home 
environment 
consisting of a Smart Lock, a Smart TV, and a set of Smart Lights, all of them interconnected via an Internet of Things (IoT) setting, and controlled, e.g., access mediated, by means of a VA. In such a scenario, the \policy, stated in Natural Language, e.g., English, restricts access to each smart device to only a subset of users, identified by their \emph{role}, e.g., \emph{Homeowner}, \emph{Partner}, etc., thus following an approach inspired by Role-Based Access Control (RBAC)~\cite{Sandhu1996}, a well-known 
access control methodology. 

For the purposes of this paper, we will consider a subset of the eXtensible Access Control Markup Language (XACML)~\cite{formal_xacml}, the \textit{de facto} standard language for authorization/access control, in an attempt to provide a consistent foundation for the \policies considered within our study, as it will be further described in Sec.~\ref{sect-methodology}. 
The \policy shown in Table~\ref{tab:sap:sample} can be expressed in a subset of XACML which
depicts a simplified syntactic structure in which \policies are composed of \emph{rules} consisting of a \emph{target}, i.e., the resource whose access is mediated, a \emph{condition} restricting under what circumstances access is to be allowed or denied, i.e., if the requester has a certain role, and a \emph{rule decision}, which states if access is either allowed or denied in case the \emph{condition} component is found to be true. 
Also, our subset of XACML includes only a single rule combining algorithm, i.e., \emph{First Applicable}, which states that the overall result of evaluating the \policy against an
access request will be the result of the first rule, starting from top to bottom, whose target is \emph{equal} to the request's target, and whose condition is evaluated as true.  

\section{Problem Statement}\label{sect-problem}

As mentioned in Sec.\ref{sect-back-related}, 
\vas may offer significant convenience by allowing users to 
interact with 
emerging technologies 
through voice or text commands. However, integrating \vas into environments that handle sensitive data and functionality introduces notable security challenges, particularly concerning the specification and enforcement of \policies, which 
are essential for 
ensuring that unauthorized individuals do not compromise security \cite{Liao2020}. The challenge lies in ensuring that \policies, as understood by the \vas, are robust enough to prevent security breaches, while maintaining usability. 
More specifically, 
in terms of \emph{effectiveness}, the management of \policies must correctly assign authorization privileges/rights, a.k.a., permissions, while avoiding the introduction of security vulnerabilities.
Conversely, in terms of \emph{efficiency}, the management of \policies must be convenient to the end-user, e.g., it avoids difficult interactions, delivering incomplete information, etc.

With that in mind, this paper aims to evaluate the effectiveness of current \vas in handling \policies across different scenarios, providing insights into their limitations and proposing new strategies for improving the secure management of sensitive data and functionality, such that future \vas can be designed to manage security while maintaining ease-of-use.    
Concretely, we consider the following research questions: 

\begin{itemize}
    \item \questionone

    \item \questiontwo
    
\end{itemize}

To further illustrate our problem statement, please consider the \policy described as a part of Sec.~\ref{sect-back-related}, also featured in Table~\ref{tab:sap:sample}. 
%
%
In such a scenario, \vas should be not only able to understand and enumerate the aforementioned \policy, but also to correctly answer to natural language questions representing access requests, which will subsequently trigger the evaluation of the \policy as described in Sec.~\ref{sect-back-related}. Such a task has interesting \emph{usability} and \emph{security} implications. For instance, access requests similar to the following: 
%
    \emph{Can Kids watch TV?}
%
may
certainly have a considerable impact in the overall usability of the whole Smart Home if they are processed incorrectly by the VA, e.g., Lights not working for the \emph{Homeowner}, which represents a noticeable inconvenience. However, the security implications may not be relevant in this case, as access to any of those two devices may not be considered as highly \emph{sensitive} for physical security purposes.
On the other hand, questions similar to the following: 
%
    \emph{Can Visitors manipulate the Smart Lock?}
%
can 
have non-trivial security consequences, as allowing for \emph{Visitors} to access a security sensitive device such as a Smart Lock can certainly represent a serious risk. \vas not evaluating \policies correctly, and therefore, granting unintended access as a result, may introduce a serious security vulnerability, which could be later exploited by malicious parties.

\begin{table}[]
\caption{Approaches for Encoding \policies for Virtual Assistants (Smart Homes)}
\label{tab:sap:sample}
\centering
\renewcommand{\arraystretch}{1.1}
\setlength{\tabcolsep}{3pt}
\begin{tabular}{p{3cm} p{4.8cm}}
\hline
Approach                                                                                  & \multicolumn{1}{c}{\policy}                                                                                                                                                                         \\ \hline
XACML \textbf{(Formal)}                                                                   & \begin{tabular}[c]{@{}l@{}}<policy, CA=First-Applicable>\\ <rule result=Allow> \\ <target>Lock</target>\\ <cond.>role=Homeowner</cond.>\\ </rule>\\ <rule result=Deny> ...</rule>\\ </policy>\end{tabular}                             \\
Informal \textbf{(INF)}                                                                   & \begin{tabular}[c]{@{}l@{}}\\ Only homeowners and partners are \\ allowed to use the Lock, everybody \\ is allowed to use the TV,  everybody\\  is allowed to use the Lights. \\ Everything else is denied.\end{tabular} \\
\begin{tabular}[c]{@{}l@{}}\\Modified Informal\\ \textbf{(Mod-INF)}\end{tabular}            & \begin{tabular}[c]{@{}l@{}}\\ Partners cannot use the lock,\\ but visitors can use the lock now. \\ \end{tabular}                                                                                                        \\
\begin{tabular}[c]{@{}l@{}}\\Semi-Formal \\ \textbf{(SEMI-UNROLL)}\end{tabular}             & \begin{tabular}[c]{@{}l@{}}\\ 1. Homeowner can access Lock, \\ 2. Homeowners can access Lights,\\ ...10. Deny access to everything else.\end{tabular}                                                                    \\
\begin{tabular}[c]{@{}l@{}}\\Modified Semi-Formal\\ \textbf{(Mod-SEMI-UNROLL)}\end{tabular} & \begin{tabular}[c]{@{}l@{}}\\ Remove Rule:\\ Partner can access Lock. \\ Create New Rule: \\Visitor can access Lock.\\ \end{tabular}                                                                                        \\
\begin{tabular}[c]{@{}l@{}}Semi-Formal-\\Rule-Based \\ \textbf{(SEMI-RULE)}\end{tabular}               & \begin{tabular}[c]{@{}l@{}}\\ 1. if role = homeowner, \\ Lock Access = allowed \\ 2. if role = partner, \\ Lock Access = allowed \\ ... 5. Deny access to everything else.\end{tabular}                                  \\
\begin{tabular}[c]{@{}l@{}}\\Modified Semi-Formal-\\Rule-Based\\ \textbf{(Mod-SEMI-RULE)}\end{tabular}   & \begin{tabular}[c]{@{}l@{}}\\ if role = visitor: lock access = allowed,\\ if role = partner: lock access = denied.\\ \end{tabular}                                                                                      \\ \hline
\end{tabular}
\end{table}

\section{Methodology}\label{sect-methodology}

To address our research questions, we conducted an exploratory study assessing the capabilities of four publicly available, general-purpose \vas:
OpenAI ChatGPT \footnote{https://openai.com/chatgpt/} (Version GPT-4o),
Google Gemini\footnote{https://gemini.google.com/app} (Version 2024.09.04),
Microsoft Copilot\footnote{https://copilot.microsoft.com/} (Version 10.28), and 
High-Flyer Deepseek\footnote{https://www.deepseek.com} (Version 2025.01.20)
on tasks related to the specification and evaluation of a series of \policies. The study began by selecting VAs that have gained significant popularity due to their recent advancements, accessibility, and relevance in general-purpose AI tasks. Next, we established the evaluation domains, providing a rationale for their selection. 
We then 
developed a series of 
\policies, offering insight into the different policy formats used in the evaluation. Subsequently, we outline the methods for conducting interactive \va sessions, distinguishing between \emph{Contextual} and \emph{Non-Contextual} methods for prompting. Finally, we analyze \va interactions, concluding the experimental phase with an evaluation of the \policies application within each \va.

\subsection{Selecting \vas}\label{selecting-vas}
ChatGPT, Gemini, Copilot, and Deepseek were chosen due to their broad user-base and frequent use, indicating that each VAs has likely accumulated knowledge from a wide variety of user interactions over its lifetime. These assistants are likely to have exposure to security-related questions and concepts making each of these \vas ideal candidates for evaluating, despite not being exclusively tailored to security tasks. 

\subsubsection{OpenAI ChatGPT}
ChatGPT, developed by OpenAI, is one of the most widely used conversational AI models. The latest version, GPT-4o, is designed to handle a broad range of tasks, including answering questions. ChatGPT has been trained on a vast dataset, enabling it to generate detailed and coherent responses across various domains. The model employs Reinforcement Learning from Human Feedback (RLHF), where developers provide desired outputs to guide the model, enhancing the structure of responses ideal for this study \cite{chatGPT}. While ChatGPT does not learn from real-time user interactions, its extensive pre-trained knowledge base includes information about 
access control models, though inconsistencies or challenges may arise due to incorrect yet plausible responses \cite{chatGPT}.

\subsubsection{Google Gemini}
Gemini, developed by Google DeepMind, is a virtual assistant that integrates advanced reasoning with internet retrieval, providing an advantage when handling up-to-date queries. It leverages Google's powerful machine learning models, which have been trained on diverse datasets across various domains \cite{Gemini}. Its multi-modal processing capabilities enable it to analyze and adapt to various formats, including those in this study. However, its reliance on web-sourced data introduces potential inconsistencies, as it may provide responses based on publicly available but unverified information \cite{Gemini}.

\subsubsection{Microsoft Copilot}
Copilot, developed by Microsoft, is a generative AI designed to assist with daily tasks across various platforms, including Office and GitHub, though this study evaluated with the general-use version to maintain consistency \cite{Copilot}. Built on the ChatGPT 4o model, Copilot is optimized for productivity and contains integration with Microsoft's security and compliance tools, making it particularly relevant for tasks involving access control and security policy interpretation in various environments \cite{copilotAI}. Although primarily focused on other uses, Copilot was included to evaluate security-handling capabilities, offering insights into its application beyond general productivity tasks. 

\subsubsection{High-Flyer Deepseek}
Deepseek, developed by High-Flyer, is an emerging \va with a focus on reasoning and structured data processing, using a Mixture-of-Experts (MoE) language model \cite{deepseek}. While it is less documented compared to the other \vas in this study, its emphasis on logic makes it an interesting candidate for evaluating security policy interpretation \cite{deepseek}. Deepseek may be able to process structured access control rules more effectively than general-purpose assistants, but its performance on natural language-based security policies is less certain. Given its relatively recent development, it is unclear how well it aligns with practices in cybersecurity or whether it introduces reasoning inconsistencies when handling access control scenarios.

\subsection{Determining Application Domains}\label{subsect-methodology-sap-spec}
We selected three application domains where \vas could significantly enhance user experience through natural language interactions. These include Smart Homes, where \vas can manage IoT devices for seamless control~\cite{edu2020smart}, Smart Cars, where \vas can assist with tasks like navigation and comfort settings~\cite{guanetti2018control, arena2020overview}, and EHRs, where \vas can support quick and accurate information retrieval in critical situations~\cite{evans2016electronic, kumah2018electronic}.

\subsubsection{Electronic Health Records Domain}
EHR systems are well-established and have become a fundamental part of modern healthcare practices. They are primarily used by healthcare professionals to store, retrieve, and update patient records, making access control an integral component to ensure the protection of private information. Most EHR systems today rely on graphical user interfaces (GUIs), although voice-based systems are becoming integrated for tasks such as searching patient records or dictating notes \cite{kumah2018electronic}.  Accuracy is another critical component in EHR systems, due to the severe or life-threatening consequences that could occur due to errors in these systems. While response times can be important, it may not be as urgent unless an emergency situation arises, where it will then become crucial for immediate access. Beyond that, EHR systems need to maintain an effective management system, to not only protect privacy and maintain regulations, but to improve the overall efficiency and reliability of healthcare delivery. Previous solutions include RBAC, which is widely used in this domain, as it ensures that different roles (e.g. doctor, nurse) have appropriate levels of access to sensitive medical data. This study explores EHRs due to the widespread adoption and the critical nature of accurate information access.  

\subsubsection{Smart Home Domain}
Smart Home technologies are becoming increasingly common, though they are not yet universally adopted in all households. These systems primarily use GUIs, but there is an increasing preference for voice control to enhance ease of use \cite{edu2020smart}. While accuracy in controlling these devices is integral, minor errors are typically not disruptive, unless they relate to security devices (e.g., cameras, locks). Slow responses, however, can reduce the overall effectiveness of the system, as users expect quick responses and actions from the systems. While management is significant, poor management implementation will cause a smart home to quickly lose effectiveness, reliability, and security. Similarly to EHRs, RBAC is also a model that could be implemented, though issues with complexities may arise due to the specific needs of users (e.g., home owners, family members) and the specific devices (e.g. thermostats, lights). Therefore, the smart home domain was selected due to the growing popularity and the potential streamlining that VAs can bring to the domain. 

\subsubsection{Smart Car Domain}
Connected Automated Vehicles (CAVs) are an emerging technology, that rely on voice control for tasks such as navigation or adjusting in-car settings (e.g. AC, radio). While some cars may have visual systems (i.e., multimedia receivers), the use of GUIs within the CAV context is not practical due to safety concerns. In terms of accuracy, critical vehicle functions could have serious consequences if errors were to occur due to incorrect policy applications. Response time is another essential component, as delays in control in a real-time environment could lead to safety risks beyond those in the CAVs. All of these components need to be effectively handled by a robust management system, ensuring that security protocols are followed to keep the system reliable and safe. Though RBAC is a possible solution, this model may not be ideal due to necessary flexibility that stems from the various users (e.g. driver, passenger) and dynamic contexts (e.g., autonomous vs. manual driving). The addition of the smart car domain was valuable for this study as the domain is still in its development stages, allowing an evaluation on the \vas ability to handle \policies, with little to no background knowledge. 

\subsection{Establishing \policies}\label{establishing-policies}
In generating format-specific \policies, this study employs a method that produced multiple structures of instructions from the same policy. This approach was used to evaluate \va capabilities across different formats, ranging from an informal, natural language format to a structured, formal code-like approach. Specifically, three formats were used: informal (natural language), semi-formal (concise statement-based), and semi-formal-rule-based (structured, code-like). Detailed descriptions of each \policy format, along with the steps taken to produce the \policies are provided in the following sections. Table~\ref{tab:sap:sample}, as well as  Tables \ref{tab:sap:sample-cars} and \ref{tab:sap:sample-ehrs} (listed as a part of Appendix A), provide a full listing of the \policies. 

\subsubsection{Informal (INF)}
Developed as an initial format, is intended to exercise the capabilities of VAs for handling moderately descriptive \policies in natural language. It contains different statement explaining the \policy rules, which in turn are composed of descriptions of roles, protected resources (targets), and quantity pronouns, i.e., everybody. 

\subsubsection{Formal (XACML)}
Developed from the INF format, leverages the subset of XACML in an effort to provide a formalization of each \policy from which other formats can be developed from. This format was not used directly in the procedures described here.

\subsubsection{Modified Informal (Mod-INF)}
Developed from the INF format, the modified informal structure is intended to exercise the capabilities of VAs for handling reduced descriptions of \policies in natural language. It contains a much brief statement describing the \policy rules using the same constructs as before.

\subsubsection{Semi-Formal (SEMI-UNROLL)}
Developed from the XACML format by syntactically \emph{unrolling} each XACML rule into a natural language one, in an effort to assess the capabilities of VAs for understanding \policy that are described as sequences of enumerated statements composed of a limited sequence of language constructs.

\subsubsection{Modified Semi-Formal(Mod-SEMI-UNROLL)}
Developed from the SEMI-UNROLL format, adds and/or removes certain \policy rules from the original set, in an effort to assess the capabilities of VAs for handling dynamic updates in \policies, which may constitute changes in the authorization decisions when the \policy is re-evaluated with respect to the one in the SEMI-UNROLL format. 

\subsubsection{Semi-Formal-Rule-Based (SEMI-RULE)}
Developed from the XACML format, provides a shorter version of the unrolling with respect to SEMI-UNROLL using a stricter natural language syntax in a rule form, in an effort to provide a more concise description of \policies for evaluation purposes.

\subsubsection{Modified Formal (Mod-SEMI-RULE)}
Finally, this format was developed from the SEMI-RULE format by \emph{compressing} several rules into a shorter statement, in an effort to assess the capabilities of VAs for handling succinct \policy rule-based descriptions that may lack extended explanations.

\subsection{Constructing Inquiries} \label{constructed-inquiries}
To assess the ability of the VAs to interpret and apply \policies, we designed a set of structured questions aimed at evaluating each VAs capabilities and limitations, which are visible in Table~\ref{table-questions}. The questions were constructed to test two primary aspects of the VA performance: policy retrieval accuracy, which measures how well the VA extracts and applies explicit rules from a given policy, and logical consistency, which evaluates whether the VA demonstrates any underlying rationale 
rather than 
arbitrary responses.

Each of these questions were developed to be straightforward and directly answerable based on the provided \policy within each of the application domains. For example, questions such as \emph{"Can partners manipulate the Lock?"}, or \emph{"Can visitors watch TV?"}, were used to determine whether the VA accurately retrieved and applied the \policies rules. In addition to the \policies rules, we tested the VAs ability to handle implicit reasoning and ambiguous cases by including questions where the policy did not directly state an answer. For instance, if a policy specified that \emph{" Only Admin Staff is allowed to access Personal Identifiable Information (PII) Data."}, we prompted the VA inquiring if nurses could access PII Data. These cases required the VA to infer whether such roles fell under the definition of "admin staff" based on the given context. 

While responses were primarily recorded as \emph{true} or \emph{false}, additional notes were taken for instances where a VA struggled to provide clear reasoning or justification for its answer. The results were analyzed based on correctness and reasoning quality. Correctness was determined by comparing the VAs response to the explicit policy statement, while reasoning quality was noted in cases where the VA deviated from the correct answer or provided an unclear justification. 

\begin{table*}[t]
\caption{Access Request Questions for Three Different \policy Scenarios}
\label{table-questions}
\begin{tabular}{lclclc}
\hline
\multicolumn{2}{c}{\textbf{Smart Homes}}                                                                                & \multicolumn{2}{c}{\textbf{Smart Cars}}                                                                              & \multicolumn{2}{c}{\textbf{EHRs}}                                                                                           \\
\multicolumn{1}{c}{\textbf{ID}} & \textbf{Question}                                                                     & \multicolumn{1}{c}{\textbf{ID}} & \textbf{Question}                                                                  & \multicolumn{1}{c}{\textbf{ID}} & \textbf{Question}                                                                         \\ \hline
Q-SM-1                          & \textit{\begin{tabular}[c]{@{}c@{}}Can Visitors manipulate \\ the Lock?\end{tabular}} & \cellcolor[HTML]{FFFFFF}Q-SC-1  & \textit{\begin{tabular}[c]{@{}c@{}}Can Drivers give\\ Directions?\end{tabular}}    & \cellcolor[HTML]{FFFFFF}Q-EHR-1 & \textit{\begin{tabular}[c]{@{}c@{}}Can Physicians access \\ Medical Data?\end{tabular}}   \\
Q-SM-2                          & \textit{Can Visitors watch TV?}                                                       & \cellcolor[HTML]{FFFFFF}Q-SC-2  & \textit{\begin{tabular}[c]{@{}c@{}}Can Co-Pilots give \\ Directions?\end{tabular}} & \cellcolor[HTML]{FFFFFF}Q-EHR-2 & \textit{\begin{tabular}[c]{@{}c@{}}Can Staff access \\ Medical Data?\end{tabular}}        \\
Q-SM-3                          & \textit{\begin{tabular}[c]{@{}c@{}}Can Homeowner turn on \\ the Lights?\end{tabular}} & \cellcolor[HTML]{FFFFFF}Q-SC-3  & \textit{\begin{tabular}[c]{@{}c@{}}Can Passengers set \\ the Radio?\end{tabular}}  & \cellcolor[HTML]{FFFFFF}Q-EHR-3 & \textit{\begin{tabular}[c]{@{}c@{}}Can Nurses access \\ PII-Data?\end{tabular}}           \\
Q-SM-4                          & \textit{\begin{tabular}[c]{@{}c@{}}Can Partner manipulate \\ the Lock?\end{tabular}}  & \cellcolor[HTML]{FFFFFF}Q-SC-4  & \textit{\begin{tabular}[c]{@{}c@{}}Can Drivers set \\ the Radio?\end{tabular}}     & \cellcolor[HTML]{FFFFFF}Q-EHR-4 & \textit{\begin{tabular}[c]{@{}c@{}}Can Staff access \\ PII-Data?\end{tabular}}            \\
Q-SM-5                          & \textit{\begin{tabular}[c]{@{}c@{}}Can Homeowner access \\ a Meter?\end{tabular}}     & \cellcolor[HTML]{FFFFFF}Q-SC-5  & \textit{\begin{tabular}[c]{@{}c@{}}Can Kids give \\ Directions?\end{tabular}}      & \cellcolor[HTML]{FFFFFF}Q-EHR-5 & \textit{\begin{tabular}[c]{@{}c@{}}Can First Responders access \\ PII-Data?\end{tabular}} \\ \hline
\end{tabular}
\end{table*}

\subsection{Conducting Interactive Sessions} \label{interative-sessions}
To evaluate the capabilities of \vas in interpreting and applying \policies, we designed two distinct methods: \emph{Contextual} and \emph{Non-Contextual} methods. These approaches were developed to assess whether prior contextual information influences a \vas ability to retrieve and apply policy information accurately. The Contextual Method simulated scenarios where VAs have domain-specific knowledge before policy evaluation, while the Non-Contextual Method assesses how well a \va performs when given policies without any prior contextual grounding. By comparing these methods, we aim to determine the extent to which contextual grounding affects policy comprehension and retrieval accuracy.

\begin{figure}[t]
    \centering
    \includegraphics[width=\columnwidth]{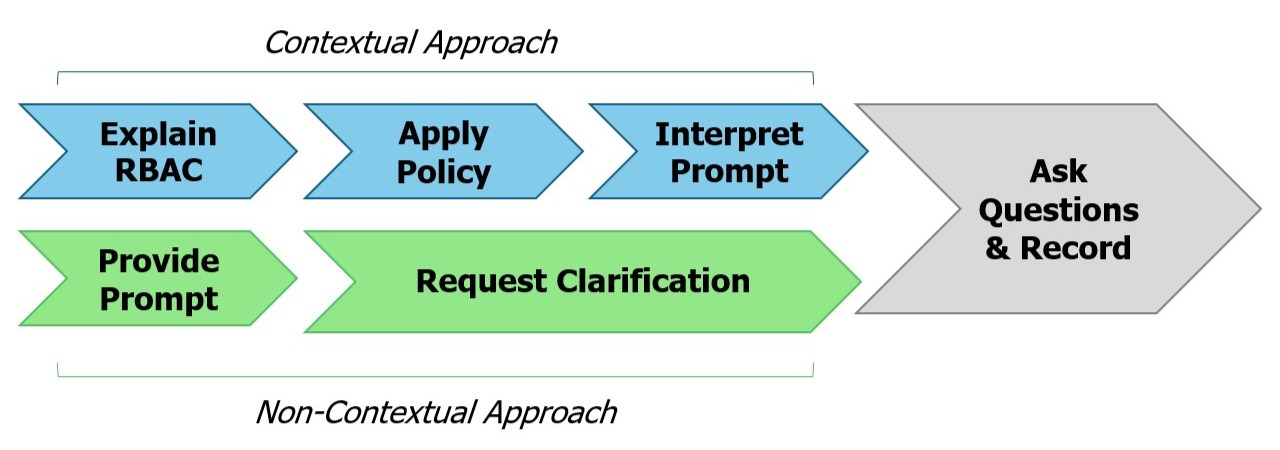} 
    \caption{Contextual and Non-Contextual Approach Steps for \va Interaction Described in Sec.~\ref{subsect-methodology-contextual} and Sec.~\ref{subsect-methodology-non-contextual}.}
    \label{fig:approach}
    \vspace{-10pt}
\end{figure}

The interactive sessions were conducted during two periods: the first two weeks of September 2024 and the first two weeks of March 2025. Three research team members with moderate training in \policy theory and \va-related technologies interacted independently with each \va. The session structure is shown in Figure \ref{fig:approach}. Afterward, two additional team members -- one expert in \policies and one expert in \va technologies -- collaborated to compile and compare the results against ground truth, ultimately deriving the recommendations to be presented in Sec.~\ref{sect-results}.

\subsubsection{Contextual Method}\label{subsect-methodology-contextual}
In the Contextual method, \vas were provided with background information before being presented with \policy-related queries. This step aimed to simulate a scenario where the \va had access to domain-specific knowledge that could aid in reasoning about policies. The process included:

    \begin{enumerate}
        \item Prompting the \va to explain RBAC, based on its pre-existing knowledge and training data, including definitions and examples.
        \item Asking the \va about application domains (e.g., Smart Homes or Smart Cars) and relevant security considerations.
        \item Requesting the \va to generate a sample RBAC-like \policy, from instructions, specific to the given domain.
        \item The generated \policy was evaluated using predefined questions from Table \ref{table-questions}, with responses compared against ground truth values.
        \item Each response was recorded, including any assumptions, inconsistencies, or deviations.
        \item After recording the results, provide the \va with the modified rules and repeat the process.
        \item Once the modified rules were recorded, the session was closed and memory was cleared.
    \end{enumerate}

\subsubsection{Non-Contextual Method}\label{subsect-methodology-non-contextual}
In contrast, the 
Non-Contextual Method
involved directly presenting the \policy within a fresh chat session, without any background information, ensuring that responses relied solely on its pre-existing knowledge. The VA received the policy statement and was immediately tested on its ability to apply the rules correctly. The procedure followed:
    \begin{enumerate}
        \item Provide the \policy to the \va within in a new chat session.
        \item Evaluate the \va by prompting the predefined questions from Table \ref{table-questions}, with responses compared against ground truth values.
        \item Record responses and in some cases, \vas responded with apologies when questioned about the incorrect responses ~\cite{wester2024ai}.
        \item After recording the results, provide the \va with the modified rules and repeat the process.
        \item Once the modified rules were recorded, the session was closed and memory was cleared.
    \end{enumerate}

\section{Results}\label{sect-results}
The evaluation of Gemini, ChatGPT, Copilot, and Deepseek revealed distinct performance characteristics across various domains. The VAs were assessed in five key areas: \textbf{Effectiveness} (Sec.~\ref{subsect-results-effectiveness}), 
\textbf{Consistency} (Sec~\ref{subsect-results-consistency}), 
\textbf{Perception} (Sec.~\ref{subsect-results-perception}), 
\textbf{Reasoning} (Sec.~\ref{subsect-results-reasoning}), and 
\textbf{Usability} (Sec.~\ref{subsect-results-usability}). 
While each VA had strengths and weaknesses, all struggled with inference-based tasks and understanding default permissions, highlighting areas for improvement, though longer interactions generally led to performance gains. Across domains, VAs faced challenges in accurately retrieving and applying \policies, with some performed better in specific formats. 
Sections \ref{subsect-results-effectiveness} to \ref{subsect-results-usability} are then focused on providing answers to \textbf{RQ1}, whereas Sec.~\ref{subsect-results-recommendations} provides a set of interesting \textbf{Recommendations} to answer \textbf{RQ2}. 

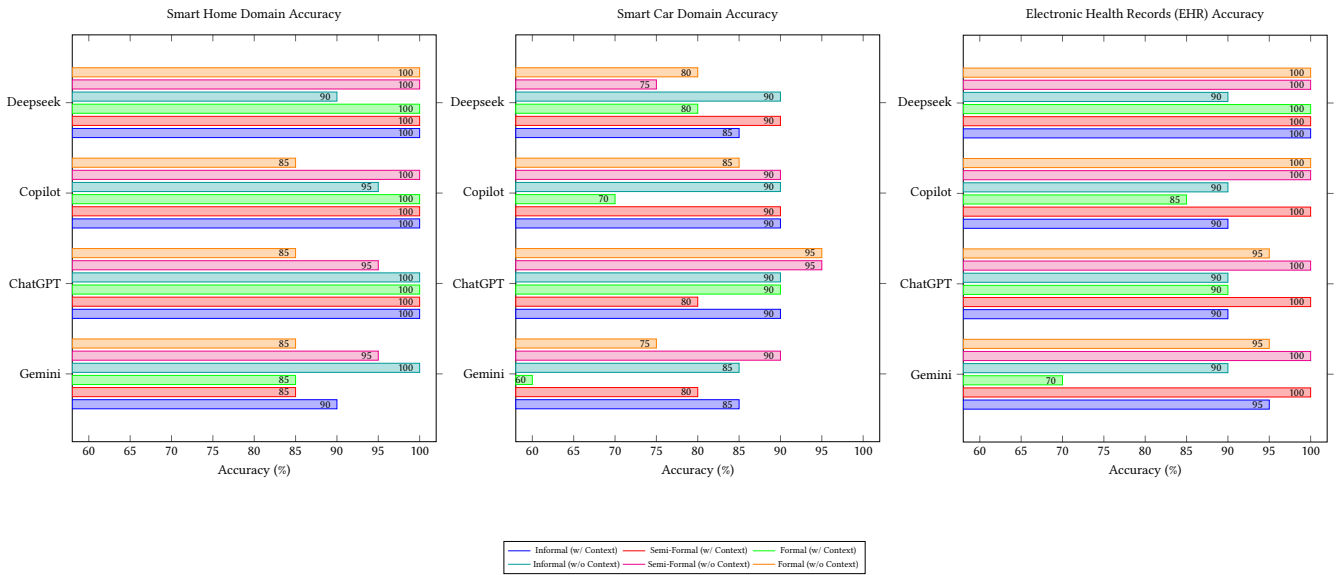
\begin{figure*}[t]
    \centering
    \resizebox{1\textwidth}{!}{
\begin{minipage}{0.5\textwidth}
    \centering
    \resizebox{\textwidth}{!}{\begin{tikzpicture}
    \begin{axis}[
         xbar,
        width=10cm, 
        height=11cm, 
        bar width=6pt,
        symbolic y coords={Gemini, ChatGPT, Copilot, Deepseek},
        ytick=data,
        xmin=60, xmax=100,
        xlabel={Accuracy (\%)},
        enlarge y limits=0.25, 
        enlarge x limits=0.05,
        title={Smart Home Domain Accuracy},
        nodes near coords,
        every node near coord/.append style={
            font=\tiny, 
            anchor=east, 
            xshift=-1pt, 
            font=\footnotesize, 
            text=black 
        },
        cycle list={
            {blue, fill=blue!30}, 
            {red, fill=red!30}, 
            {green, fill=green!30}, 
            {cyan, fill=cyan!30}, 
            {magenta, fill=magenta!30},
            {orange, fill=orange!30}
        }
    ]
    
    \addplot coordinates {(90,Gemini) (100,ChatGPT) (100,Copilot) (100,Deepseek)};
    \addplot coordinates {(85,Gemini) (100,ChatGPT) (100,Copilot) (100,Deepseek)};
    \addplot coordinates {(85,Gemini) (100,ChatGPT) (100,Copilot) (100,Deepseek)};
    
    \addplot coordinates {(100,Gemini) (100,ChatGPT) (95,Copilot) (90,Deepseek)};;
    \addplot coordinates {(95,Gemini) (95,ChatGPT) (100,Copilot) (100,Deepseek)};
    \addplot coordinates {(85,Gemini) (85,ChatGPT) (85,Copilot) (100,Deepseek)};;
    
    \end{axis}
\end{tikzpicture}} 
    \label{fig:premodification}
\end{minipage}\hfill
\begin{minipage}{0.5\textwidth}
    \centering
    \resizebox{\textwidth}{!}{\begin{tikzpicture}
    \begin{axis}[
        xbar,
        width=10cm, 
        height=11cm, 
        bar width=6pt,
        symbolic y coords={Gemini, ChatGPT, Copilot, Deepseek},
        ytick=data,
        xmin=60, xmax=100,
        xlabel={Accuracy (\%)},
        enlarge y limits=0.25, 
        enlarge x limits=0.05,
        title={Smart Car Domain Accuracy},
        nodes near coords,
        every node near coord/.append style={
            font=\tiny, 
            anchor=east, 
            xshift=-1pt, 
            font=\footnotesize, 
            text=black 
        },
        cycle list={
            {blue, fill=blue!30}, 
            {red, fill=red!30}, 
            {green, fill=green!30}, 
            {cyan, fill=cyan!30}, 
            {magenta, fill=magenta!30},
            {orange, fill=orange!30}
        }
    ]
    
    \addplot coordinates {(85,Gemini) (90,ChatGPT) (90,Copilot) (85,Deepseek)};
    \addplot coordinates {(80,Gemini) (80,ChatGPT) (90,Copilot) (90,Deepseek)};
    \addplot coordinates {(60,Gemini) (90,ChatGPT) (70,Copilot) (80,Deepseek)};
    
    \addplot coordinates {(85,Gemini) (90,ChatGPT) (90,Copilot) (90,Deepseek)};
    \addplot coordinates {(90,Gemini) (95,ChatGPT) (90,Copilot) (75,Deepseek)};
    \addplot coordinates {(75,Gemini) (95,ChatGPT) (85,Copilot) (80,Deepseek)};
    
    \end{axis}
\end{tikzpicture}}
    \label{fig:postmodification}
\end{minipage}\hfill
\vspace{1em}
\begin{minipage}{0.5\textwidth}
    \centering
    \resizebox{\textwidth}{!}{\begin{tikzpicture}
    \begin{axis}[
      xbar,
        width=10cm, 
        height=11cm, 
        bar width=6pt,
        symbolic y coords={Gemini, ChatGPT, Copilot, Deepseek},
        ytick=data,
        xmin=60, xmax=100,
        xlabel={Accuracy (\%)},
        enlarge y limits=0.25, 
        enlarge x limits=0.05,
        title={Electronic Health Records (EHR) Accuracy},
        nodes near coords,
        every node near coord/.append style={
            font=\tiny, 
            anchor=east, 
            xshift=-1pt, 
            font=\footnotesize, 
            text=black 
        },
        cycle list={
            {blue, fill=blue!30}, 
            {red, fill=red!30}, 
            {green, fill=green!30}, 
            {cyan, fill=cyan!30}, 
            {magenta, fill=magenta!30},
            {orange, fill=orange!30}
        }
    ]
    
    \addplot coordinates {(95,Gemini) (90,ChatGPT) (90,Copilot) (100,Deepseek)};
    \addplot coordinates {(100,Gemini) (100,ChatGPT) (100,Copilot) (100,Deepseek)};
    \addplot coordinates {(70,Gemini) (90,ChatGPT) (85,Copilot) (100,Deepseek)};
    
    \addplot coordinates {(90,Gemini) (90,ChatGPT) (90,Copilot) (90,Deepseek)};
    \addplot coordinates {(100,Gemini) (100,ChatGPT) (100,Copilot) (100,Deepseek)};
    \addplot coordinates {(95,Gemini) (95,ChatGPT) (100,Copilot) (100,Deepseek)};
    
    \end{axis}
\end{tikzpicture}} 
    \label{fig:overall}
\end{minipage}\hfill}
\resizebox{0.25\textwidth}{!}{
\begin{minipage}{0.5\textwidth}
    \centering
    \begin{tikzpicture}
        \begin{axis}[
            axis line style={draw=none},
            tick style={draw=none},
            xtick=\empty,
            ytick=\empty,
            enlargelimits=false,
            scale only axis=true,
            width=5cm,
            height=1cm,
            legend columns=3,
            legend style={
                    at={(0,0)},
                    anchor=north,
                    font=\footnotesize
                },
            cycle list={
                {blue, fill=blue!30}, 
                {red, fill=red!30}, 
                {green, fill=green!30}, 
                {cyan, fill=cyan!30}, 
                {magenta, fill=magenta!30},
                {orange, fill=orange!30}
            }
        ]
            \addplot[blue] coordinates {(0,0)};
            \addplot[red] coordinates {(0,0)};
            \addplot[green] coordinates {(0,0)};
            \addplot[cyan] coordinates {(0,0)};
            \addplot[magenta] coordinates {(0,0)};
            \addplot[orange] coordinates {(0,0)};
            \legend{Informal (w/ Context), Semi-Formal (w/ Context), Formal (w/ Context), Informal (w/o Context), Semi-Formal (w/o Context), Formal (w/o Context)}%
        \end{axis}
    \end{tikzpicture}
    \label{fig:legend}
\end{minipage}}
\caption{\centering Overall Session Accuracy for Specific Domains (Smart Home, Smart Car, Electronic Health Records) - With and Without Context.}
\label{fig:domains}
\end{figure*}
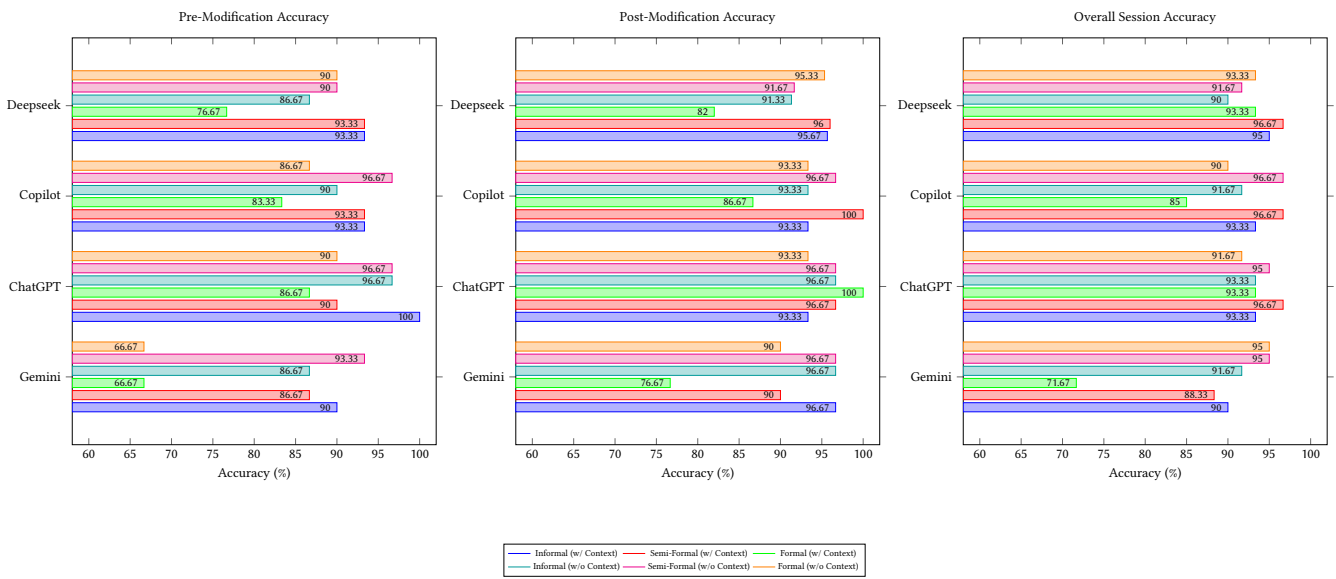
\begin{figure*}[t]\label{overallsession}
    \centering
    \resizebox{1\textwidth}{!}{
\begin{minipage}{0.5\textwidth}
    \centering
    \resizebox{\textwidth}{!}{\begin{tikzpicture}
    \begin{axis}[
        xbar,
        width=10cm, 
        height=11cm, 
        bar width=6pt,
        symbolic y coords={Gemini, ChatGPT, Copilot, Deepseek},
        ytick=data,
        xmin=60, xmax=100,
        xlabel={Accuracy (\%)},
        enlarge y limits=0.25, 
        enlarge x limits=0.05,
        title={Pre-Modification Accuracy},
        nodes near coords,
        every node near coord/.append style={
            font=\tiny, 
            anchor=east, 
            xshift=-1pt, 
            font=\footnotesize, 
            text=black 
        },
        cycle list={
            {blue, fill=blue!30}, 
            {red, fill=red!30}, 
            {green, fill=green!30}, 
            {cyan, fill=cyan!30}, 
            {magenta, fill=magenta!30},
            {orange, fill=orange!30}
        }
    ]
    
    \addplot coordinates {(90.00,Gemini) (100.00,ChatGPT) (93.33,Copilot)(93.33,Deepseek)};
    \addplot coordinates {(86.67,Gemini) (90.00,ChatGPT) (93.33,Copilot) (93.33,Deepseek)};
    \addplot coordinates {(66.67,Gemini) (86.67,ChatGPT) (83.33,Copilot) (76.67,Deepseek)};
    
    \addplot coordinates {(86.67,Gemini) (96.67,ChatGPT) (90.00,Copilot) (86.67,Deepseek)};
    \addplot coordinates {(93.33,Gemini) (96.67,ChatGPT) (96.67,Copilot) (90,Deepseek)};
    \addplot coordinates {(66.67,Gemini) (90.00,ChatGPT) (86.67,Copilot) (90,Deepseek)};
    
    \end{axis}
\end{tikzpicture}} 
    \label{fig:premodification}
\end{minipage}\hfill
\begin{minipage}{0.5\textwidth}
    \centering
    \resizebox{\textwidth}{!}{\begin{tikzpicture}\label{overall-session-chart}
    \begin{axis}[
         xbar,
        width=10cm, 
        height=11cm, 
        bar width=6pt,
        symbolic y coords={Gemini, ChatGPT, Copilot, Deepseek},
        ytick=data,
        xmin=60, xmax=100,
        xlabel={Accuracy (\%)},
        enlarge y limits=0.25, 
        enlarge x limits=0.05,
        title={Post-Modification Accuracy},
        nodes near coords,
        every node near coord/.append style={
            font=\tiny, 
            anchor=east, 
            xshift=-1pt, 
            font=\footnotesize, 
            text=black 
        },
        cycle list={
            {blue, fill=blue!30}, 
            {red, fill=red!30}, 
            {green, fill=green!30}, 
            {cyan, fill=cyan!30}, 
            {magenta, fill=magenta!30},
            {orange, fill=orange!30}
        }
    ]
    
    \addplot coordinates {(96.67,Gemini) (93.33,ChatGPT) (93.33,Copilot) (95.67,Deepseek)};
    \addplot coordinates {(90.00,Gemini) (96.67,ChatGPT) (100.00,Copilot) (96.00,Deepseek)};
    \addplot coordinates {(76.67,Gemini) (100.00,ChatGPT) (86.67,Copilot) (82.00,Deepseek)};
    
    \addplot coordinates {(96.67,Gemini) (96.67,ChatGPT) (93.33,Copilot) (91.33,Deepseek)};
    \addplot coordinates {(96.67,Gemini) (96.67,ChatGPT) (96.67,Copilot) (91.67,Deepseek)};
    \addplot coordinates {(90.00,Gemini) (93.33,ChatGPT) (93.33,Copilot) (95.33,Deepseek)};
    
    \end{axis}
\end{tikzpicture}}
    \label{fig:postmodification}
\end{minipage}\hfill
\vspace{1em}
\begin{minipage}{0.5\textwidth}
    \centering
    \resizebox{\textwidth}{!}{\begin{tikzpicture}
    \begin{axis}[
 xbar,
        width=10cm, 
        height=11cm, 
        bar width=6pt,
        symbolic y coords={Gemini, ChatGPT, Copilot, Deepseek},
        ytick=data,
        xmin=60, xmax=100,
        xlabel={Accuracy (\%)},
        enlarge y limits=0.25, 
        enlarge x limits=0.05,
        title={Overall Session Accuracy},
        nodes near coords,
        every node near coord/.append style={
            font=\tiny, 
            anchor=east, 
            xshift=-1pt, 
            font=\footnotesize, 
            text=black 
        },
        cycle list={
            {blue, fill=blue!30}, 
            {red, fill=red!30}, 
            {green, fill=green!30}, 
            {cyan, fill=cyan!30}, 
            {magenta, fill=magenta!30},
            {orange, fill=orange!30}
        }
    ]
    
    \addplot coordinates {(90.00,Gemini) (93.33,ChatGPT) (93.33,Copilot) (95,Deepseek)};
    \addplot coordinates {(88.33,Gemini) (96.67,ChatGPT) (96.67,Copilot) (96.67,Deepseek)};
    \addplot coordinates {(71.67,Gemini) (93.33,ChatGPT) (85.00,Copilot) (93.33,Deepseek)};
    
    \addplot coordinates {(91.67,Gemini) (93.33,ChatGPT) (91.67,Copilot) (90,Deepseek)};
    \addplot coordinates {(95.00,Gemini) (95.00,ChatGPT) (96.67,Copilot) (91.67,Deepseek)};
    \addplot coordinates {(95.00,Gemini) (91.67,ChatGPT) (90.00,Copilot) (93.33,Deepseek)};
    
    \end{axis}
\end{tikzpicture}} 
    \label{fig:overall}
\end{minipage}\hfill}
\resizebox{0.25\textwidth}{!}{
\begin{minipage}{0.5\textwidth}
    \centering
    \begin{tikzpicture}
        \begin{axis}[
            axis line style={draw=none},
            tick style={draw=none},
            xtick=\empty,
            ytick=\empty,
            enlargelimits=false,
            scale only axis=true,
            width=5cm,
            height=1cm,
            legend columns=3,
            legend style={
                    at={(0,0)},
                    anchor=north,
                    font=\footnotesize
                },
            cycle list={
                {blue, fill=blue!30}, 
                {red, fill=red!30}, 
                {green, fill=green!30}, 
                {cyan, fill=cyan!30}, 
                {magenta, fill=magenta!30},
                {orange, fill=orange!30}
            }
        ]
            \addplot[blue] coordinates {(0,0)};
            \addplot[red] coordinates {(0,0)};
            \addplot[green] coordinates {(0,0)};
            \addplot[cyan] coordinates {(0,0)};
            \addplot[magenta] coordinates {(0,0)};
            \addplot[orange] coordinates {(0,0)};
            \legend{Informal (w/ Context), Semi-Formal (w/ Context), Formal (w/ Context), Informal (w/o Context), Semi-Formal (w/o Context), Formal (w/o Context)}%
        \end{axis}
    \end{tikzpicture}
    \label{fig:legend}
\end{minipage}}
\caption{\centering Overall Session Accuracy, Pre-Modified Session Accuracy, \& Post-Modified Session Accuracy for All Scenarios(Smart Home, Smart Car, Electronic Health Records) - With and Without Context.}
\label{fig:overall}
\end{figure*}

\subsection{Effectiveness}\label{subsect-results-effectiveness}
Effectiveness measures how well a \va can apply \policies correctly with pre-defined rules or ground truth values. An important and primary function of \vas in this context is to make decisions based on security policies, as incorrect decisions can cause the system and users to be at risk. This study aims to evaluate the overall effectiveness of \vas, which directly impacts 
their security.  

ChatGPT performs well across all formats and methods. Accuracy ranges from 91.67\% to 96.67\%, with the lowest at 80\% in the Smart Car domain and the highest at 100\% across various others, as seen in Figure ~\ref{fig:domains}. This consistent accuracy positions ChatGPT as a strong contender for future VA advancements.

Gemini shows promise, specifically in usability, but lags behind other \vas in overall accuracy, with scores ranging from 71.67\% to 95\%. Its lowest score is 60\% in the Smart Car domain, while it achieves 100\% in other domains. These mixed results suggest areas for improvement, especially in the pre-training process.

Copilot, built on ChatGPT, ranges from 85\% to 96.67\% in accuracy, but drops to 70\% in the Smart Car domain. While it performs similarly to ChatGPT, it may not be as robust for handling \policies, especially when compared to other specialized \vas like Deepseek.

Deepseek, an emerging \va, demonstrates strong performance with accuracies ranging from 90\% to 96.6\%. This reasoning-based model allows it to produce results comparable to ChatGPT, with a low of 75\% and a high of 100\% across various domains. Deepseek and ChatGPT are among the top performers for applying \policies, making them contenders for VA and security applications.

\subsection{Consistency}\label{subsect-results-consistency}
Consistency is a critical measure of \vas reliability in recalling and applying \policies in repeated interactions. Inconsistent behavior can lead to confusion or misinterpretation, particularly in scenarios where access control rules must be applied uniformly. In real-world applications, security policies must be consistently enforced to maintain integrity and prevent vulnerabilities. Notably, longer interaction sessions generally led to improved consistency, suggesting that extended engagement allows \vas to refine their responses and better adhere to previously established policies. As shown in Figure~\ref{fig:overall}, the improvement in post-modification accuracy highlights the benefits of restating rules, which enables the system to better retrieve and apply information.

ChatGPT, like most \vas, showed a steady increase in accuracy as sessions progressed, even when re-prompted with information that adjusted the original rules. Starting with an average accuracy of 93.34\% in the pre-modification phase, ChatGPT achieved an average accuracy of 96.11\% in the post-modification phase. 

Gemini, though scoring low in pre-modifi\-ca\-tion, with an average accuracy of 81.67\%, was able to redeem its score through the longer interactive sessions. Once the rules were restated and modified, Gemini was able to increase its average accuracy to 91.11\%, though that is one of the lowest of the post-modification scores when compared. 

Copilot, though able to improve its average accuracy in the post-modifi\-ca\-tion phase, had the lowest pre-modification score at an average of 90.56\%. This suggests that longer interactions may be necessary for information retention. After rule modifications and restatements, Copilot’s accuracy increased by three percent, reaching an average of 93.89\%.

Deepseek, despite being an emerging \va, surprisingly had the second-lowest pre-modification average accuracy at 88.33\%, suggesting that longer sessions may be necessary for establishing a clearer \policy. Similar to Copilot, Deepseek improved in the post-modification phase, increasing its average accuracy to 92\%.

\subsection{Perception}\label{subsect-results-perception}
Perception, in the context of this study, refers to how the instructions are formatted and the accuracy of interpretation from each \va. Each format, ranging from informal, natural language to semi-formal, rule-based, code-like structures, will indicate varying level of understanding that are dependent on the pre-training process. Some \vas were able to produce results across all formats, while others excelled or struggled with specific formats. Perception aims to understand the best format to provide \policies 
for 
higher accuracy. 

Each \va (ChatGPT, Gemini, Copilot, and Deepseek) achieved relatively high accuracy when handling the \emph{Informal} format, with scores ranging from 86.67\% to 100\%. This indicates that most VAs effectively process natural language requests. While the majority performed well, as seen in Table~\ref{table-results-no-accuracy-smart-home}, Deepseek consistently produced lower results in this format, which may be attributed to its pre-training process and how it interprets informal inputs. Since \policies will likely be provided in natural language from users, ensuring strong performance in this format is crucial for real-world applications.

While the \vas performed well in the natural language (\emph{Informal}) format, the \emph{Semi-Formal} format yielded slightly higher scores, ranging from 88.33\% to 100\%. This difference may be attributed to the structured nature of \policies in this format, where rules are explicitly stated with clear roles and conditions. Notably, each VA either matched or outperformed its informal format score, suggesting that the semi-formal structure may be more effective for providing \vas with specific rules to apply accurately.

While the \vas performed well with the \emph{Informal} and \emph{Semi-Formal} policy formats, the \emph{Semi-Formal-Rule-Based} format resulted in a notable decline in accuracy. Scores ranged from 66.67\% to 100\%, with this wide gap likely attributed to the code-like structure of the \policy. Despite being trained on large-scale datasets, these models are primarily optimized for natural language processing and may have less exposure to structured, rule-based formats. However, ChatGPT and Deepseek performed significantly better, in comparison, in this format likely due to differences in the training processes.

\subsection{Reasoning}\label{subsect-results-reasoning}
Reasoning evaluates the logical structure of responses, particularly when it comes to justifying decisions based on \policies. The ability to reason effectively in dynamic environments is essential where new or complex policies may need to be applied. 

Each VA produced structured responses, highlighting either the potential for future reasoning capabilities or the current implementation of logical reasoning, as seen with Deepseek. While Deepseek benefits from built-in reasoning, other \vas struggled with inference-based questions. For instance, when asked, \emph{"Can homeowners access a meter?"} a question relying on the \policy statement \emph{"everything else is denied"}, the VA must infer that the meter falls under "everything else" and is therefore not accessible. Since the meter is not explicitly mentioned, some VAs failed to recognize this inference and incorrectly marked it as accessible.

Though inference was a consistent struggle across all domains, all \vas provided responses with relatively similar structures, with only a few exceeding the basic format. As expected, both Copilot and ChatGPT followed the same response pattern, which includes an immediate answer (i.e., \emph{"No, the Homeowner cannot access a Meter"}), followed by a justification (i.e., \emph{"Your policy states that everything else is denied unless explicitly allowed"}). These responses were generally brief and accurate, but in undetermined answer cases
an
additional
apology 
was included, 
(i.e., \emph{"I apologize for not understanding your policy"}), 
which inspired the title of this paper. 

In comparison, Gemini uses a similar structure, but adds an additional component to the end, which includes a summary of the \policy. This approach allows the \va to restate the \policy without user prompting, possibly reinforcing the \policies. However, as shown in Figures 2 and 3, reinforcing incorrect policies can negatively impact the accuracy of the \va (i.e., \emph{"We cannot determine if a homeowner can access a meter"}). 

Deepseek, following a similar structure to Gemini, includes an additional component in its responses by providing a suggestion. These suggestions vary depending on the \policy, but may include recommendations for resolving denied access issues or improving the overall model (i.e., \emph{"Example modifications"}).

\subsection{Usability} \label{subsect-results-usability}
Usability measures how easy and efficient it is for users to interact with a VA. This includes the response time, quality of interaction, and any potential issues such as blocking or delays from the \va. While each \va offers varying levels of response times and message limits, each VA evaluation is necessary to understand the underlying potential solutions and limitations. 

ChatGPT demonstrated consistent usability across various contexts. Its response times were relatively quick, averaging approximately 3.15 seconds across different scenarios. While response generation was immediate, we observed that ChatGPT imposed a message limit, particularly when multiple messages were sent within a one-hour period. 
We found that 
limit was triggered after approximately sixty to sixty-five messages were sent, by notifying the user that they \emph{"reached our [ChatGPT] limit of messages per hour."} Notably, this threshold was recorded in September 2024, and with each new version release, there appears to be a corresponding increase in the number of messages allowed within the hour.

Gemini, another strong contender in the VA environment, demonstrated quick response times across various contexts. Like ChatGPT, Gemini delivered responses promptly, averaging approximately 3.21 seconds across the differing formats. Although response times were fast, Gemini also imposed a message limit when multiple messages were sent within a one-hour window. This message limit was triggered after approximately one hundred messages were sent, with users being notified that \emph{“Gemini is on a break.”} 

Copilot, a VA built on ChatGPT, performed slightly better in response time but has a much smaller message threshold. It generates responses to prompts, whether questions or instructions, relatively quickly, averaging approximately 2.88 seconds. 
However, 
Copilot has a much smaller message limit, restricting entire chat sessions to only thirty messages before requiring the user to open a new chat. Additionally, Copilot imposes a message limit within the one-hour window similar to ChatGPT, typically around the sixty-message mark, informing users that it is \emph{"time for a new topic"} or that the user has \emph{"reached the limit of messages in one hour"}. 

Deepseek, an emerging VA, incorporates reasoning into its base model, resulting in longer response times, as observed in the study. Although Deepseek provides detailed and clear responses, it averages a generation time of approximately 14 seconds, which is significantly longer than other VAs. This extended response time is likely correlated with the reasoning aspect of the model. Additionally, Deepseek's message threshold is smaller but not implemented in the same way as Copilot, Gemini, or ChatGPT. Deepseek imposes a temporary barrier after approximately twenty messages, requiring users to pause for five to ten minutes before they can send messages again by notifying the user that there is a \emph{"temporary server error"}. This temporary restriction occurs three times, after which the user is blocked for the remainder of the hour by notifying that the user has \emph{"reached the message limit for the hour"}. 

\subsection{Recommendations}\label{subsect-results-recommendations}
Each of these previously mentioned components (effectiveness, consistency, perception, reasoning, usability) highlight the strengths and limitations of each \va. The results from these sections point to a key underlying issue of a lack of inference and logical structure. At the time of testing, most models lacked a dedicated reasoning framework. As a result, incorrect responses often stemmed from prompts requiring inference whether it was related to specific devices or roles. Beyond that, pre-training information also plays a part in the accuracy of these models, as domain results are representative of how widespread the domain is.

\subsubsection{Implementing Reasoning into \vas}
In order to address the lack of reasoning in models, one notable solution is to implement a form of reasoning. As demonstrated by Deepseek, this approach can significantly enhance the quality of responses. However, it may lead to a trade-off in usability, specifically in terms of response time. To avoid compromising usability of \vas, it is crucial to incorporate reasoning without negatively impacting response speed. Deepseek, while incorporating reasoning, experiences much longer response times compared to other \vas. Although the models previously evaluated lacked reasoning integration, ChatGPT has since released a reasoning model which adds approximately four-to-six seconds to the average response time \cite{teamai}.
This makes its response time closer to Deepseek's but still retains a reasonable balance between logic and speed. These changes could improve the effectiveness, consistency, and clarity of each \va, allowing the \va to be utilized across various domains. 

\subsubsection{Importance of Pre-Training Information}
To enhance the quality of \vas responses, the training process plays a crucial role. Analyzing the results across various domains reveals that the EHR domain consistently achieved high accuracy. These scores may be attributed to the widespread use of access control within the healthcare field, which provides a solid foundation for training within this domain. On the other hand, the Smart Home domain scored decently across all \vas, but exhibited inference errors. Since smart homes are less widespread and not as extensively documented as EHRs, the lack of supporting documentation in the training process may point to the importance of background information. Specifically, for a \va to be integrated into a Smart Home domain, there may be necessary tailoring process to introduce specific access control concepts and applications. In contrast, Smart Cars face an even greater challenge. Not only is there limited documentation available when compared to EHRs, but the documentation on security, particularly access control, is scarce. This lack of security-related training materials may contribute to challenges \vas face when operating within the Smart Car domain. Each of these domains contribute to the overall conclusion that there is not enough supporting documentation of emerging technologies, as \vas are often updated with information at a later time. 

\subsubsection{Enhancing Real-Time Adaptability}
A major limitation in current \vas is their inability to learn in real-time. Most VAs rely on pre-training data, meaning they cannot adapt to new information or refine their understanding based on user interactions. This is particularly problematic in domains that evolve quickly or lack comprehensive documentation, such as emerging technologies like Smart Cars. In these dynamic fields, the inability of \vas to update their knowledge can result in outdated or inaccurate responses, limiting their usefulness. Introducing real-time learning capabilities would enable \vas to continually improve and adjust their responses during interactions. This enhancement would make the \vas more dynamic and adaptable, offering more relevant responses while significantly improving accuracy and utility. Though, it is important to note that learning in real-time could allow unverified and incorrect information to be taught to the model, therefore, a real-time model would require a meticulous implementation. 

\section{Limitations} \label{sect-limitations}
While this study provides valuable insights into the limitations and strengths of VAs, several limitations should be considered when interpreting the results. 

\subsubsection{Simplification of \policies}
This study, while exploring various domains, had  \policies that were based on simple case scenarios. While the \policies were able to showcase underlying issues within the \vas, the policies were not exhaustive (e.g. no corner cases and/or complicated policies with combining algorithms). In order to address this limitation, a future study needs to be conducted to explore richer and more complicated \policies that require multi-step decision-making. Further exploration into complex cases and \policies would showcase the importance of reasoning within \vas, and how the usability will be impacted. 

\subsubsection{Narrow Range of Perspectives}
Though the results were produced from few researchers, each of these researches that interacted with the \vas contained varying levels of expertise: expert, mid-level, and novice. In comparison, the end-users who may interact with the \vas will likely not have the same levels of expertise in security or access control methods. Therefore, it is worth noting that a future usability study will need to be conducted with participants who have limited security experience, to gauge how policies are understood and managed from an outside perspective.  

\subsubsection{Limitations of \vas}
For this study, only a handful of the most popular \vas were chosen, as they would likely have the most human interaction, in turn, having the most training. Though these \vas would have the benefit of extensive training, only general-purpose, commercially available, non-specialized \vas were considered. In light of this limitation, future work for this study includes leveraging insight from the \vas strengths and weaknesses to develop a custom \va that focuses on handling \policies 
consistently. 

\section{Conclusions and Future Work}\label{sect-conclusions}

In this paper, we have presented an exploratory study to evaluate
the effectiveness of the publicly-available \vas in managing \policies for a set of representative domain case scenarios. 
%
%
Our results indicate the need to further customize \vas to effectively handle \policies to prevent security vulnerabilities, while not infringing on the user experience. Overall, the \vas demonstrated varying results, where some excelled in formal approaches but declined in performance for informal ones. Each of the \vas showcased above-average comprehension in at least one of formatted \policy approaches, but also highlighted some possible limitations in other ones. 
%
Although these results are encouraging, 
their power is limited to the well-known RBAC approach. 
In the future, 
we plan to extend our investigation to other authorization approaches, i.e., Attribute-Based Access Control (ABAC) \cite{IEEEexample:NIST_ABAC},  
which may be better suited for the representative domain scenarios we introduced for our study. 
Finally, we also plan to further implement our proposed recommendations in a series of custom-made \vas, which can effectively leverage these experience for enhanced security and usability performance.


\begin{acks}
This work was partially supported by the National Science Foundation (NSF) under Grants No. 2232911 and No. 2131263, and
by the CAHSI-Google Institutional Research Program, sponsored by Google Inc. and the Computing Alliance of Hispanic-Serving Institutions (CAHSI), a National Science Foundation (NSF) INCLUDES Alliance.
\end{acks}

\bibliographystyle{ACM-Reference-Format}
\bibliography{references}

\clearpage

\section{Appendix A: \policiesfull and Sample Results} \label{appendix-saps}
\begin{table}[h]
\caption{Approaches for Encoding \policies for Virtual Assistants (Smart Cars)}
\label{tab:sap:sample-cars}
\begin{tabular}{@{}cl@{}}
\hline
Approach                                                                                  & \multicolumn{1}{c}{\policy}                                                                                                                                                                                                                                               \\ \hline
\begin{tabular}[c]{@{}c@{}}XACML \\ (Formal)\end{tabular}                                 & \begin{tabular}[c]{@{}l@{}}<policy, CA=First-Applicable>\\  <rule result=Deny> \\ <target>Directions</target> \\ <cond.>role=Co-Pilot</cond.>\\ </rule>\\  </policy>\end{tabular} \\                                                                                                                                   \\
\begin{tabular}[c]{@{}c@{}}Informal \\ \textbf{(INF)}\end{tabular}                        & \textit{\begin{tabular}[c]{@{}l@{}}Only drivers should be allowed to \\ give driving directions to the \\ smart car. Only drivers and co-pilots\\  should be allowed to modify smart \\ car settings. Other passengers \\ should be only allowed to change \\ the radio settings.\end{tabular}} \\ \\
\begin{tabular}[c]{@{}c@{}}Modified Informal\\ \textbf{(Mod-INF)}\end{tabular}            & \textit{\begin{tabular}[c]{@{}l@{}}Kids can't give directions, \\ but co-pilots can.\end{tabular}}       \\                                                                                                                                                                                       \\
\begin{tabular}[c]{@{}c@{}}Semi-Formal \\ \textbf{(SEMI-UNROLL)}\end{tabular}             & \begin{tabular}[c]{@{}l@{}}1. Co-Pilots cannot give Directions,\\ 2. Passengers cannot give Directions,\\ ... 4. Everything else is Allowed.\end{tabular}   \\                                                                                                                                    \\
\begin{tabular}[c]{@{}c@{}}Modified Semi-Formal\\ \textbf{(Mod-SEMI-UNROLL)}\end{tabular} & \begin{tabular}[c]{@{}l@{}}Create new Rule: \\ Kids cannot give Directions; \\ Remove Rule: \\ Co-Pilots cannot give Directions\end{tabular}        \\                                                                                                                                            \\
\begin{tabular}[c]{@{}c@{}}Semi-Formal-\\Rule-Based \\ \textbf{(SEMI-RULE)}\end{tabular}               & \begin{tabular}[c]{@{}l@{}}if Role = Co-Pilot: \\ Directions = denied, \\ if Role = Passenger: \\ Settings = denied\\  \& Directions = denied, \\ Default Radio \\ \& Directions \\ \& Settings = approved\end{tabular}      \\                                                                   \\
\begin{tabular}[c]{@{}c@{}}Modified Semi-Formal-\\Rule-Based\\ \textbf{(Mod-SEMI-RULE)}\end{tabular}   & \begin{tabular}[c]{@{}l@{}}if Role = Kids: \\ Directions = denied, \\ if Role = Co-Pilot: \\ Directions = approved\end{tabular}    \\                                                                                                                                                             \\ \hline
\end{tabular}
\end{table}
\begin{table}[h]
\caption{Approaches for Encoding \policies for Virtual Assistants (EHRs)}
\label{tab:sap:sample-ehrs}
\begin{tabular}{@{}cl@{}}
\hline
Approach                                                                                              & \multicolumn{1}{c}{\policy}                                                                                                                                                                                     \\ \hline
\begin{tabular}[c]{@{}c@{}}XACML \\ (Formal)\end{tabular}                                             & \begin{tabular}[c]{@{}l@{}}<policy, CA=First-Applicable>\\ <rule result=Allow>\\ <target>Medical</target>\\ <cond.>role=Physician</cond.>\\ </rule> \\ <rule result=Allow>...</rule>\\ </policy>\end{tabular} \\                                              \\
\begin{tabular}[c]{@{}c@{}}Informal \\ \textbf{(INF)}\end{tabular}                                    & \textit{\begin{tabular}[c]{@{}l@{}}Only Physicians, Nurses, and \\ First Responders are allowed \\ to access Medical Data. \\ Only Admin Staff is allowed \\ to access Personal Identifiable \\ Information (PII) Data.\end{tabular}} \\ \\
\begin{tabular}[c]{@{}c@{}}Modified Informal\\ \textbf{(Mod-INF)}\end{tabular}                        & \textit{\begin{tabular}[c]{@{}l@{}}First Responders can now \\ access PII-Data and Staff \\ can now access Medical Data.\end{tabular}}   \\                                                                                             \\
\begin{tabular}[c]{@{}c@{}}Semi-Formal \\ \textbf{(SEMI-UNROLL)}\end{tabular}                         & \begin{tabular}[c]{@{}l@{}}1. Physicians can access Medical Data, \\ 2. Nurses can access Medical Data, \\ ... 9. Everything else is denied.\end{tabular}\\                                                                             \\
\begin{tabular}[c]{@{}c@{}}Modified Semi-Formal\\ \textbf{(Mod-SEMI-UNROLL)}\end{tabular}             & \begin{tabular}[c]{@{}l@{}}Create new Rule: \\ "First Responders can \\ access PII- Data"; \\ Remove Rule: \\ "First Responders cannot \\ access PII-Data"; \\ Create new Rule:\\ "Staff can access Medical Data"\\ \end{tabular} \\         \\
\begin{tabular}[c]{@{}c@{}}Semi-Formal-Rule-Based\\ \textbf{(SEMI-RULE)}\end{tabular}                 & \begin{tabular}[c]{@{}l@{}}if role = Staff: \\ PII-Data = Approve; \\ if role = Physician, Nurse, \\ First Responders: \\ Medical Data = Approved; \\ Default Access = Denied\end{tabular}         \\                                      \\
\begin{tabular}[c]{@{}c@{}}Modified Semi-Formal-\\ Rule-Based\\ \textbf{(Mod-SEMI-RULE)}\end{tabular} & \begin{tabular}[c]{@{}l@{}}if role = Staff, First Responders: \\ PII-Data = Approve; \\ if role = Physician, Nurse, \\ First Responders, Staff:\\ Medical Data = Approved; \\ Default Access = Denied\\ \end{tabular} \\                  \\ \hline
\end{tabular}
\end{table}
\clearpage
\begin{table*}[]
\caption{Sample Results of Evaluating Different \policies on a Series of \vas (Smart Homes)}
\label{table-results-no-accuracy-smart-home}
\begin{tabular}{cccccccccccccccccc}
\hline
\textbf{}         & \textbf{}   & \multicolumn{4}{c}{\textbf{ChatGPT}}                                                                   & \multicolumn{4}{c}{\textbf{Google Gemini}}                                                                          & \multicolumn{4}{c}{\textbf{Microsoft Copilot}}                                       & \multicolumn{4}{c}{\textbf{Deepseek}}                                                                                                                                \\
\textbf{}         & \textbf{}   & \multicolumn{2}{c}{\textbf{Context}} & \multicolumn{2}{c}{\textbf{No Context}}                         & \multicolumn{2}{c}{\textbf{Context}}                                & \multicolumn{2}{c}{\textbf{No Context}}       & \multicolumn{2}{c}{\textbf{Context}} & \multicolumn{2}{c}{\textbf{No Context}}       & \multicolumn{2}{c}{\textbf{Context}}                                               & \multicolumn{2}{c}{\textbf{No Context}}                                         \\
\textbf{Question} & \textbf{GT} & \textbf{R-1}     & \textbf{R- 2}     & \textbf{R-1}                   & \textbf{R-2}                   & \textbf{R-1} & \textbf{R-2}                                         & \textbf{R-1} & \textbf{R-2}                   & \textbf{R-1}      & \textbf{R-2}     & \textbf{R-1} & \textbf{R-2}                   & \textbf{R-1}                                      & \textbf{R-2}                   & \textbf{R-1}                                       & \textbf{R-2}               \\ \hline
\multicolumn{18}{c}{\textbf{INF Format}}                                                                                                                                                                                                                                                                                                                                                                                                                                                                                     \\ \hline
Q-SM-1            & \FALSE      & \FALSE           & \FALSE            & \FALSE                         & \FALSE                         & \FALSE       & \FALSE                                               & \FALSE       & \FALSE                         & \FALSE            & \FALSE           & \FALSE       & \FALSE                         & \multicolumn{1}{l}{\FALSE}                        & \multicolumn{1}{l}{\FALSE}     & \multicolumn{1}{l}{\FALSE}                         & \multicolumn{1}{l}{\FALSE} \\
Q-SM-2            & \TRUE       & \TRUE            & \TRUE             & \TRUE                          & \TRUE                          & \TRUE        & \TRUE                                                & \TRUE        & \TRUE                          & \TRUE             & \TRUE            & \TRUE        & \TRUE                          & \TRUE                                             & \TRUE                          & \TRUE                                              & \TRUE                      \\
Q-SM-3            & \TRUE       & \TRUE            & \TRUE             & \TRUE                          & \TRUE                          & \TRUE        & \TRUE                                                & \TRUE        & \TRUE                          & \TRUE             & \TRUE            & \TRUE        & \TRUE                          & \TRUE                                             & \TRUE                          & \TRUE                                              & \TRUE                      \\
Q-SM-4            & \TRUE       & \TRUE            & \TRUE             & \TRUE                          & \TRUE                          & \TRUE        & \TRUE                                                & \TRUE        & \TRUE                          & \TRUE             & \TRUE            & \TRUE        & \TRUE                          & \TRUE                                             & \TRUE                          & \TRUE                                              & \TRUE                      \\
Q-SM-5            & \FALSE      & \FALSE           & \FALSE            & \FALSE                         & \FALSE                         & \FALSE       & \cellcolor[HTML]{FFCCC9}\TRUE                        & \FALSE       & \FALSE                         & \FALSE            & \FALSE           & \FALSE       & \FALSE                         & \FALSE                                            & \FALSE                         & \FALSE                                             & \FALSE                     \\ \hline
\multicolumn{18}{c}{\textbf{Modified INF Format}}                                                                                                                                                                                                                                                                                                                                                                                                                                                                            \\ \hline
Q-SM-1            & \TRUE       & \TRUE            & \TRUE             & \TRUE                          & \TRUE                          & \TRUE        & \TRUE                                                & \TRUE        & \TRUE                          & \TRUE             & \TRUE            & \TRUE        & \TRUE                          & \TRUE                                             & \TRUE                          & \multicolumn{1}{l}{\cellcolor[HTML]{FFCCC9}\FALSE} & \TRUE                      \\
Q-SM-2            & \TRUE       & \TRUE            & \TRUE             & \TRUE                          & \TRUE                          & \TRUE        & \TRUE                                                & \TRUE        & \TRUE                          & \TRUE             & \TRUE            & \TRUE        & \TRUE                          & \TRUE                                             & \TRUE                          & \TRUE                                              & \TRUE                      \\
Q-SM-3            & \TRUE       & \TRUE            & \TRUE             & \TRUE                          & \TRUE                          & \TRUE        & \TRUE                                                & \TRUE        & \TRUE                          & \TRUE             & \TRUE            & \TRUE        & \TRUE                          & \TRUE                                             & \TRUE                          & \TRUE                                              & \TRUE                      \\
Q-SM-4            & \FALSE      & \FALSE           & \FALSE            & \FALSE                         & \FALSE                         & \FALSE       & \FALSE                                               & \FALSE       & \FALSE                         & \FALSE            & \FALSE           & \FALSE       & \cellcolor[HTML]{FFCCC9}\TRUE  & \FALSE                                            & \FALSE                         & \multicolumn{1}{l}{\cellcolor[HTML]{FFCCC9}\TRUE}  & \FALSE                     \\
Q-SM-5            & \FALSE      & \FALSE           & \FALSE            & \FALSE                         & \FALSE                         & \FALSE       & \cellcolor[HTML]{FFCCC9}\TRUE                        & \FALSE       & \FALSE                         & \FALSE            & \FALSE           & \FALSE       & \FALSE                         & \FALSE                                            & \FALSE                         & \FALSE                                             & \FALSE                     \\ \hline
\multicolumn{18}{c}{\textbf{SEMI-UNROLL Format}}                                                                                                                                                                                                                                                                                                                                                                                                                                                                             \\ \hline
Q-SM-1            & \FALSE      & \FALSE           & \FALSE            & \FALSE                         & \FALSE                         & \FALSE       & \FALSE                                               & \FALSE       & \FALSE                         & \FALSE            & \FALSE           & \FALSE       & \FALSE                         & \FALSE                                            & \FALSE                         & \FALSE                                             & \FALSE                     \\
Q-SM-2            & \TRUE       & \TRUE            & \TRUE             & \TRUE                          & \TRUE                          & \TRUE        & \TRUE                                                & \TRUE        & \TRUE                          & \TRUE             & \TRUE            & \TRUE        & \TRUE                          & \TRUE                                             & \TRUE                          & \TRUE                                              & \TRUE                      \\
Q-SM-3            & \TRUE       & \TRUE            & \TRUE             & \TRUE                          & \TRUE                          & \TRUE        & \TRUE                                                & \TRUE        & \TRUE                          & \TRUE             & \TRUE            & \TRUE        & \TRUE                          & \TRUE                                             & \TRUE                          & \TRUE                                              & \TRUE                      \\
Q-SM-4            & \TRUE       & \TRUE            & \TRUE             & \TRUE                          & \TRUE                          & \TRUE        & \TRUE                                                & \TRUE        & \cellcolor[HTML]{FFCCC9}\FALSE & \TRUE             & \TRUE            & \TRUE        & \TRUE                          & \TRUE                                             & \TRUE                          & \TRUE                                              & \TRUE                      \\
Q-SM-5            & \FALSE      & \FALSE           & \FALSE            & \FALSE                         & \FALSE                         & \FALSE       & \cellcolor[HTML]{FFCCC9}\TRUE                        & \FALSE       & \FALSE                         & \FALSE            & \FALSE           & \FALSE       & \FALSE                         & \FALSE                                            & \FALSE                         & \FALSE                                             & \FALSE                     \\ \hline
\multicolumn{18}{c}{\textbf{Modified SEMI-UNROLL Format}}                                                                                                                                                                                                                                                                                                                                                                                                                                                                    \\ \hline
Q-SM-1            & \TRUE       & \TRUE            & \TRUE             & \FALSE                         & \TRUE                          & \TRUE        & \TRUE                                                & \TRUE        & \TRUE                          & \TRUE             & \TRUE            & \TRUE        & \TRUE                          & \TRUE                                             & \TRUE                          & \TRUE                                              & \TRUE                      \\
Q-SM-2            & \TRUE       & \TRUE            & \TRUE             & \TRUE                          & \TRUE                          & \TRUE        & \TRUE                                                & \TRUE        & \TRUE                          & \TRUE             & \TRUE            & \TRUE        & \TRUE                          & \TRUE                                             & \TRUE                          & \TRUE                                              & \TRUE                      \\
Q-SM-3            & \TRUE       & \TRUE            & \TRUE             & \TRUE                          & \TRUE                          & \TRUE        & \TRUE                                                & \TRUE        & \TRUE                          & \TRUE             & \TRUE            & \TRUE        & \TRUE                          & \TRUE                                             & \TRUE                          & \TRUE                                              & \TRUE                      \\
Q-SM-4            & \FALSE      & \FALSE           & \FALSE            & \FALSE                         & \FALSE                         & \FALSE       & \FALSE                                               & \FALSE       & \FALSE                         & \FALSE            & \FALSE           & \FALSE       & \FALSE                         & \FALSE                                            & \FALSE                         & \FALSE                                             & \FALSE                     \\
Q-SM-5            & \FALSE      & \FALSE           & \FALSE            & \FALSE                         & \FALSE                         & \FALSE       & \cellcolor[HTML]{FFCCC9}\TRUE                        & \FALSE       & \FALSE                         & \FALSE            & \FALSE           & \FALSE       & \FALSE                         & \multicolumn{1}{l}{\cellcolor[HTML]{FFCCC9}\TRUE} & \FALSE                         & \FALSE                                             & \FALSE                     \\ \hline
\multicolumn{18}{c}{\textbf{SEMI-RULE Format}}                                                                                                                                                                                                                                                                                                                                                                                                                                                                               \\ \hline
Q-SM-1            & \FALSE      & \FALSE           & \FALSE            & \FALSE                         & \FALSE                         & \FALSE       & \FALSE                                               & \FALSE       & \FALSE                         & \FALSE            & \FALSE           & \FALSE       & \cellcolor[HTML]{FFCCC9}\TRUE  & \FALSE                                            & \FALSE                         & \FALSE                                             & \FALSE                     \\
Q-SM-2            & \TRUE       & \TRUE            & \TRUE             & \TRUE                          & \cellcolor[HTML]{FFCCC9}\FALSE & \TRUE        & \TRUE                                                & \TRUE        & \cellcolor[HTML]{FFCCC9}\FALSE & \TRUE             & \TRUE            & \TRUE        & \cellcolor[HTML]{FFCCC9}\FALSE & \cellcolor[HTML]{FFCCC9}\FALSE                    & \cellcolor[HTML]{FFCCC9}\FALSE & \TRUE                                              & \TRUE                      \\
Q-SM-3            & \TRUE       & \TRUE            & \TRUE             & \TRUE                          & \TRUE                          & \TRUE        & \TRUE                                                & \TRUE        & \TRUE                          & \TRUE             & \TRUE            & \TRUE        & \TRUE                          & \cellcolor[HTML]{FFCCC9}\FALSE                    & \cellcolor[HTML]{FFCCC9}\FALSE & \TRUE                                              & \TRUE                      \\
Q-SM-4            & \TRUE       & \TRUE            & \TRUE             & \TRUE                          & \TRUE                          & \TRUE        & \TRUE                                                & \TRUE        & \TRUE                          & \TRUE             & \TRUE            & \TRUE        & \TRUE                          & \TRUE                                             & \TRUE                          & \TRUE                                              & \TRUE                      \\
Q-SM-5            & \FALSE      & \FALSE           & \FALSE            & \FALSE                         & \FALSE                         & \FALSE       & \cellcolor[HTML]{FFCCC9}\TRUE                        & \FALSE       & \cellcolor[HTML]{FFCCC9}\TRUE  & \FALSE            & \FALSE           & \FALSE       & \FALSE                         & \FALSE                                            & \FALSE                         & \FALSE                                             & \FALSE                     \\ \hline
\multicolumn{18}{c}{\textbf{Modified SEMI-RULE Format}}                                                                                                                                                                                                                                                                                                                                                                                                                                                                      \\ \hline
Q-SM-1            & \TRUE       & \TRUE            & \TRUE             & \cellcolor[HTML]{FFCCC9}\FALSE & \TRUE                          & \TRUE        & \cellcolor[HTML]{FFCCC9}\FALSE                       & \TRUE        & \TRUE                          & \TRUE             & \TRUE            & \TRUE        & \TRUE                          & \TRUE                                             & \TRUE                          & \TRUE                                              & \TRUE                      \\
Q-SM-2            & \TRUE       & \TRUE            & \TRUE             & \TRUE                          & \cellcolor[HTML]{FFCCC9}\FALSE & \TRUE        & \TRUE                                                & \TRUE        & \TRUE                          & \TRUE             & \TRUE            & \TRUE        & \cellcolor[HTML]{FFCCC9}\FALSE & \cellcolor[HTML]{FFCCC9}\FALSE                    & \cellcolor[HTML]{FFCCC9}\FALSE & \TRUE                                              & \TRUE                      \\
Q-SM-3            & \TRUE       & \TRUE            & \TRUE             & \TRUE                          & \TRUE                          & \TRUE        & \TRUE                                                & \TRUE        & \TRUE                          & \TRUE             & \TRUE            & \TRUE        & \TRUE                          & \cellcolor[HTML]{FFCCC9}\FALSE                    & \cellcolor[HTML]{FFCCC9}\FALSE & \TRUE                                              & \TRUE                      \\
Q-SM-4            & \FALSE      & \FALSE           & \FALSE            & \FALSE                         & \FALSE                         & \FALSE       & \FALSE                                               & \FALSE       & \FALSE                         & \FALSE            & \FALSE           & \FALSE       & \FALSE                         & \FALSE                                            & \FALSE                         & \FALSE                                             & \FALSE                     \\
Q-SM-5            & \FALSE      & \FALSE           & \FALSE            & \FALSE                         & \FALSE                         & \FALSE       & \cellcolor[HTML]{FFCCC9}{\color[HTML]{000000} \TRUE} & \FALSE       & \cellcolor[HTML]{FFCCC9}\TRUE  & \FALSE            & \FALSE           & \FALSE       & \FALSE                         & \FALSE                                            & \FALSE                         & \FALSE                                             & \FALSE                     \\ \hline
\end{tabular}
\end{table*}

\end{document}